\documentclass[runningheads]{llncs}
\usepackage{graphicx}
\usepackage{tikz}
\usepackage{comment}
\usepackage{amsmath,amssymb,pifont} % define this before the line numbering.
\usepackage{color}

\usepackage[accsupp]{axessibility}  % Improves PDF readability for those with disabilities.

\usepackage{multirow}
\usepackage{booktabs}
\usepackage{hyperref}

\newsavebox\CBox
\def\textBF#1{\sbox\CBox{#1}\resizebox{\wd\CBox}{\ht\CBox}{\textbf{#1}}}

\begin{document}
% \renewcommand\thelinenumber{\color[rgb]{0.2,0.5,0.8}\normalfont\sffamily\scriptsize\arabic{linenumber}\color[rgb]{0,0,0}}
% \renewcommand\makeLineNumber {\hss\thelinenumber\ \hspace{6mm} \rlap{\hskip\textwidth\ \hspace{6.5mm}\thelinenumber}}
% \linenumbers
\pagestyle{headings}
\mainmatter
\def\ECCVSubNumber{1804}  % Insert your submission number here

\title{Real-RawVSR: Real-World Raw Video Super-Resolution with a Benchmark Dataset} % Replace with your title

% INITIAL SUBMISSION 
\begin{comment}
\titlerunning{ECCV-22 submission ID \ECCVSubNumber} 
\authorrunning{ECCV-22 submission ID \ECCVSubNumber} 
\author{Anonymous ECCV submission}
\institute{Paper ID \ECCVSubNumber}
\end{comment}
%******************

% CAMERA READY SUBMISSION
% \begin{comment}
\titlerunning{Real-RawVSR}
% If the paper title is too long for the running head, you can set
% an abbreviated paper title here
%
\author{Huanjing Yue \and
Zhiming Zhang \and
Jingyu Yang\thanks{This work was supported in part by the National Natural Science
Foundation of China under Grant 62072331. \textit{Corresponding author: Jingyu Yang.}}}
\authorrunning{Yue \textit{et al.}}
% First names are abbreviated in the running head.
% If there are more than two authors, 'et al.' is used.
%
\institute{School of Electrical and Information Engineering, Tianjin University, Tianjin, China\\
\email{\{huanjing.yue, zmzhang, yjy\}@tju.edu.cn}}
% \end{comment}
%******************
\maketitle

\begin{abstract}

In recent years, real image super-resolution (SR) has achieved promising results due to the development of SR datasets and corresponding real SR methods. In contrast, the field of real video SR is lagging behind, especially for real raw videos. Considering the superiority of raw image SR over sRGB image SR, we construct a real-world raw video SR (Real-RawVSR) dataset and propose a corresponding SR method.
We utilize two DSLR cameras and a beam-splitter to simultaneously capture low-resolution (LR) and high-resolution (HR) raw videos with $2\times$, $3\times$, and $4\times$ magnifications. There are 450 video pairs in our dataset, with scenes varying from indoor to outdoor, and  motions including camera and object movements. To our knowledge, this is the first real-world raw VSR dataset.
Since the raw video is characterized by the Bayer pattern, we propose a two-branch network, which deals with both the packed RGGB sequence and the original Bayer pattern sequence, and the two branches are complementary to each other. After going through the proposed co-alignment, interaction, fusion, and reconstruction modules, we generate the corresponding HR sRGB sequence. Experimental results demonstrate that the proposed method outperforms benchmark real and synthetic video SR methods with either raw or sRGB inputs.
\textit{Our code and dataset are available at \href{https://github.com/zmzhang1998/Real-RawVSR}{https://github.com/zmzhang1998/Real-RawVSR}.}

\end{abstract}
%------------------------------------------------------------------------
\section{Introduction}
\sloppy
\label{sec:intro}

Capturing images (videos) with a short-focus lens can enlarge the view angles by sacrificing the resolutions while capturing with a long-focus lens can increase the resolutions by sacrificing the view angles. Image (video) super-resolution (SR) is an effective way to get both wide angle and  high-resolution (HR) images (videos).   
Video SR reconstructs an HR video from a low-resolution (LR) input by exploring the spatial and temporal correlations of the input sequence. In recent years, the development of video SR has shifted from traditional model-driven to deep learning based methods~\cite{chan2021basicvsr,tian2020tdan,wang2019edvr,xue2019video}.   

The performance of these deep learning based SR methods heavily depends on the training datasets. Considering that the synthetic LR-HR datasets, such as DIV2K~\cite{agustsson2017ntire} and REDS~\cite{nah2019ntire}, cannot represent the degradation models between real captured LR images and HR images, many real SR datasets are constructed to boost the real-world SR performance. However, most of these datasets are for static LR-HR images, such as RealSR~\cite{cai2019toward} and ImagePairs~\cite{joze2020imagepairs}. Recently, Yang \textit{et al.} \cite{yang2021real} proposed the first real-world video SR dataset via capturing with a multi-camera system of iPhone 11 Pro Max. However, the parallax between the LR and HR cameras increased the difficulty for alignment and there are only $2\times$ LR-HR sequence pairs in this dataset due to the limited focal lengths of phone cameras. 

%there are obvious misalignments between the LR and HR pairs due to the parallax between the two cameras. 

On the other hand, there is a trend to utilize raw images for real-scene image (video) restoration, such as low light enhancement \cite{chen2019seeing,chen2018learning},  denoising~\cite{abdelhamed2020ntire,abdelhamed2019ntire,brooks2019unprocessing,liu2019learning,wang2020practical,yue2020supervised}, deblurring~\cite{liang2020raw}, and super-resolution~\cite{xu2019towards,zhang2019zoom}. The main reason is that raw images have wide bit depths (12 or 14 bits), \textit{i.e.,} containing the most original information, and its intensity is linear to the illumination. However, there is still little work exploring raw video SR. Liu \textit{et al.} \cite{liu2021exploit} proposed a raw video SR dataset by synthesizing  LR raw frames by downsampling from the captured HR raw frames. Even though, there is still a gap between the synthesized LR raw frames and real captured ones, which makes the SR models trained on synthesized data cannot generalize well to real scenes. 

Based on the above observations, we propose to construct a real-world raw video SR dataset to facilitate the raw VSR research. Specifically, we build a two-camera system with a beam-splitter to make sure that there is no parallax between the two cameras. In addition, we perform alignment on the captured LR-HR pairs to make them aligned. 
%By analyzing the above VSR datasets, we build a real raw VSR benchmark dataset for real-world SR tasks.
%Inspired by~\cite{joze2020imagepairs}, we build a data capture platform to collect data.
%We use two Canon cameras with optical zoom to shoot HR and LR dynamic scenes at the same time, and use a beam-splitter to ensure that there is no parallax between the two cameras.
%We refer to the method in~\cite{liu2021exploit} and use the camera with a third-party software to directly capture raw videos.
%Then we use some traditional algorithms to solve the misalignment problem caused by lens distortion and device movement offset, and preprocess and crop the data to get the final dataset.
On the other hand, the current VSR methods~\cite{chan2021basicvsr,wang2019edvr} are mostly based on sRGB frame inputs and the network design for raw sequence inputs has not been well explored. Therefore, we propose a raw VSR network tailored for raw inputs. Specifically, the raw frames are fed into the network in two forms. One is in its original Bayer pattern and the other is in the packed sub-frame version, namely that RGGB pixels are packed into four channels. The features from the two branches are co-aligned, interacted, and fused together to reconstruct the HR sRGB frame.  
%We use raw data as input and pack it into 4 channels, \ie RGGB, to deeply mine the depth features of raw input.
%After bidirectional progressive alignment module, reconstruction module, and upsampling module, the final HR sRGB video sequence is obtained.
%Our experiments prove that the proposed network is more conducive to using raw data to achieve state-of-the-art results.
In brief, our contributions can be summarized as follows.
\begin{itemize}
\item[$\bullet$] We construct the first aligned raw VSR dataset for real scenes, which contains LR-HR pairs for $2\times$, $3\times$, and $4\times$ magnification in both raw and sRGB domains. By utilizing a beam splitter in our capturing system, we obtain LR-HR pairs without parallax. There are totally 450 video pairs and each video contains about 150 frames.
\item[$\bullet$] We propose a novel raw VSR network by utilizing the raw frames in terms of the original Bayer pattern and its corresponding packed sub-frame pattern. Specifically, we propose co-alignment, interaction, and fusion modules to take advantage of the complementary information from the two branches.
%The proposed two branch based solution outperforms single branch based solution with only slightly increase of   
%that can directly process LR raw data and produce superior results, which is better than the state-of-the-art VSR method of base and sRGB.
\item[$\bullet$] We introduce a simple but effective color correction method (\textit{i.e.,} channel-based correction), which is beneficial for training with image pairs having color differences. Experimental results demonstrate that our method outperforms benchmark VSR methods in both sRGB and raw domains. 
%Our dataset not only contains raw-pair data, but also paired LR-HR sRGB video sequences, which provides a benchmark dataset for the subsequent VSR research of real-world data.
\end{itemize}
%------------------------------------------------------------------------
\section{Related Work}
\label{sec:related}
%------------------------------------------------------------------------
\subsection{Image and Video SR Datasets}
{\bf Image SR datasets.} The early image SR datasets usually synthesize LR images from the captured HR ones via bicubic downsampling, such as DIV2K dataset~\cite{agustsson2017ntire}. 
Considering the domain gap between synthesized and real captured LR images, many real-world SR datasets are constructed. For example, the City100~\cite{chen2019camera} and RealSR~\cite{cai2019toward} datasets, which are captured with different focal length cameras, contain LR-HR pairs in sRGB domain. Zhang~\textit{et al.} claimed that using sRGB images to train the SR model is inferior to that trained by raw data~\cite{zhang2019zoom}. Therefore, 
they constructed the first SR-Raw dataset for real-world computational zoom. Meanwhile, Xu~\textit{et al.} constructed a synthesized raw image dataset for raw image SR \cite{xu2019towards}. Hereafter,
the ImagePairs dataset~\cite{joze2020imagepairs} is constructed by introducing a beam splitter into the capturing system, which enables them to capture a much larger dataset with LR-HR pairs in both raw and sRGB domains. These datasets have greatly promoted the performance of real image SR and laid the foundation for the construction of VSR datasets for real scenes. 

\textbf{Video SR Datasets.} Similar to the development of image SR dataset, the video dataset is also shifted from the synthesized ones (such as REDS~\cite{nah2019ntire} and Vimeo-90k~\cite{xue2019video}) to real captured ones (such as RealVSR~\cite{yang2021real} and BurstSR dataset~\footnote{Since burst image SR is similar to video SR, we present them here other than in the image SR.}~\cite{bhat2021deep}), from sRGB domain \cite{xue2019video,yang2021real} to raw domain \cite{bhat2021deep,liu2021exploit}. The RealVSR \cite{yang2021real} dataset is constructed by capturing with two different focal length cameras in iPhone 11 Pro Max and the DoubleTake App. Since the focal lengths are limited for phone cameras, there are only $2\times$ LR-HR sequence pairs in this dataset. Recently, RealBasicVSR~\cite{chan2021investigating} built a VideoLQ dataset to assess the generalize ability of real-world VSR methods. Since there are no ground truths for these videos, this dataset cannot be used for supervised training. 

Inspired by the success of raw image SR, 
Bhat \textit{et al.} constructed a BurstSR dataset~\cite{bhat2021deep} in the raw domain by capturing the burst LR raw images with a phone camera and the HR sRGB images with a DSLR camera. Liu \textit{et al.} constructed the RawVD dataset~\cite{liu2021exploit} for videos, which synthesized the LR raw sequences  from the captured HR raw sequences via a degradation model. However, as demonstrated in \cite{bhat2021deep}, a network
trained with synthetic data is expected to have suboptimal performance when applied to real images. Therefore, we propose to construct a Real-RawVSR dataset by capturing real raw sequences with both short and long focal length cameras for different scaling factors, thus providing a real benchmark for raw VSR model training and evaluation. 
%------------------------------------------------------------------------
\subsection{Image and Video SR Methods}
\textbf{SR Methods for Synthesized Data.}
In the literature, most SR methods are designed based on the synthesized LR-HR pairs. For image SR, most works explore efficient modules to explore spatial correlations, such as the residual channel attention block in RCAN \cite{zhang2018image}, the holistic attention block in HAN \cite{niu2020single}. For video SR, both spatial and temporal correlations are essential for SR performance. Therefore, many methods focus on the alignment strategy, such as the optical flow based  \cite{caballero2017real,kappeler2016video,xue2019video} and the deformable convolution \cite{dai2017deformable} based, \textit{e.g.}  TDAN \cite{tian2020tdan}, EDVR \cite{wang2019edvr}. Recently, BasicVSR~\cite{chan2021basicvsr} and its enhanced versions , \textit{i.e.,} IconVSR~\cite{chan2021basicvsr} and BasicVSR++ \cite{chan2021basicvsr++} have achieved superior SR performance by combining forward and backward bidirectional propagation information and optical flow based feature alignment. Hereafter, Zhou~\textit{et al.} proposed an effective iterative alignment algorithm and an efficient adaptive reweighting strategy to better utilize the temporal correlations \cite{zhou2021revisiting}.
 
\textbf{SR Methods for Real Captured Data.}
Different from synthesized LR-HR pairs, there are usually spatial misalignment, color mismatching, and intensity variance in the real captured LR-HR pairs. Therefore, the SR methods for real data focus on dealing with these misalignments. Zhang~\textit{et al.} introduced the contextual bilateral loss to deal with the spatial misalignment~\cite{zhang2019zoom}, and Cai~\textit{et al.} proposed a Laplacian pyramid based kernel prediction network since the real degradation kernels are naturally non-uniform~\cite{cai2019toward}. Besides, the NTIRE challenge on real-world image SR further boosts the SR performance~\cite{cai2019ntire,lugmayr2020ntire}. Compared with real image SR, there is a few research on real VSR. RealVSR~\cite{yang2021real} proposed a Laplacian pyramid based loss to deal with the misalignment and color differences between the LR-HR frames. Considering that in-the-wild degradations could be exaggerated during temporal propagation, RealBasicVSR~\cite{chan2021investigating} proposed a pre-cleaning module to reduce noise and artifacts prior to temporal propagation. 

\textbf{SR Methods for Raw Images and Videos.}
The above methods are generally designed for sRGB images. For raw input SR, the network needs to simultaneously deal with both ISP and SR tasks. The work in~\cite{zhang2019zoom} directly maps the raw input to an sRGB output via a ResNet. Different from it, Xu~\textit{et al.} proposed a dual CNN, where one branch is used for structure reconstruction and the other branch is for color restoration with the LR sRGB image as guidance~\cite{xu2019towards}. Following it, the RawVSR method~\cite{liu2021exploit}  also utilizes two branches for both detail and color reconstruction. However, the raw LR frames are synthesized. 

To our knowledge, there is still no work exploring real-world raw VSR methods and the network design for raw sequence input has not been well explored. In this work, we propose a two-branch interaction network tailored for raw sequence inputs and propose co-alignment, interaction, and fusion modules to explore the complementary information between the two branches. 

\section{Real-RawVSR Dataset Construction}
\label{sec:dataset}

\begin{figure}[t]
  \centering
  \setlength{\abovecaptionskip}{0.cm}
   \includegraphics[width=0.92\textwidth]{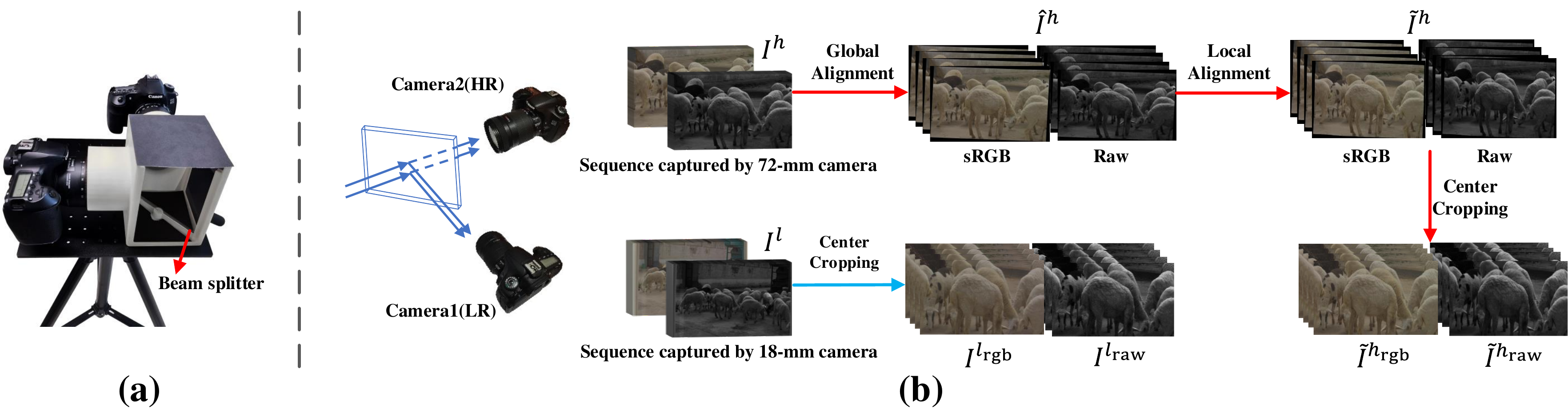}
   \caption{The capturing hardware (a) and our coarse to fine alignment pipeline to generate aligned LR-HR pairs (b).}
   \label{fig:fig1}
%   \vspace{-0.4cm}
\end{figure}

{\bf Hardware Design.} 
Capturing LR-HR image pairs with short-long focal lengths are common settings for real image SR. This can be easily realized for static scene by capturing with the same camera \cite{zhang2019zoom}. For dynamic LR-HR video capturing, we need to utilize two cameras with different focal lengths. However, this will inevitably bring parallax problems caused by different shooting positions. Inspired by~\cite{jiang2019learning,joze2020imagepairs}, which utilizes a beam splitter to divide the incident light into two light beams with a brightness ratio of 1:1, we also utilize this strategy, as shown in Fig.~\ref{fig:fig1} (a). In order to capture LR-HR frame pairs with different ratios, we utilize the DSLR camera with an 18-135mm zoom lens instead of the mobile phone cameras. Therefore, a large beam splitter is expected to cover the lens of DSLR cameras. To this end, we utilize a large and cheap beam splitter with reflectance coating and antireflection coating, instead of a small and expensive beam splitter cube. In order to avoid the influence of natural light from other directions, we design and print a 3D model box to hold the beam splitter. In this way, the two cameras can receive natural light from the same viewpoint. The size of the beam splitter is $150\times150\times1 (\text{mm}^3)$, which is enough to cover the camera lens. We put the camera and beam splitter box on an optical plate, which is installed on a tripod, to improve its stability.

%------------------------------------------------------------------------
%------------------------------------------------------------------------
{\bf Data Collection.} We use two Canon 60D cameras upgraded with a third-party software Magic Lantern \footnote{https://magiclantern.fm/} to capture raw videos in Magic Lantern Video (MLV) format. To keep the cameras in sync, we use an infrared remote control to signal both cameras to capture at the same time. During capturing, we keep the ISO of the two cameras ranging from 100 to 1600 to avoid noise, and the exposure time ranges from 1/400s to 1/31s to capture both slow and fast motions. All the other settings are set to default values to simulate real capture scenarios. Then we use the MlRawViewer\footnote{https://bitbucket.org/baldand/mlrawviewer/src/master/} software to process the MLV video to obtain the corresponding sRGB frames and raw frames in the DNG format. For each scene, we capture a short video with six seconds and the frame rate is 25 FPS, namely that each video contains approximately 150 frames in both raw and sRGB formats. 

%------------------------------------------------------------------------
\begin{figure}[t]
  \centering
  \setlength{\abovecaptionskip}{0.cm}
   \includegraphics[width=0.92\linewidth]{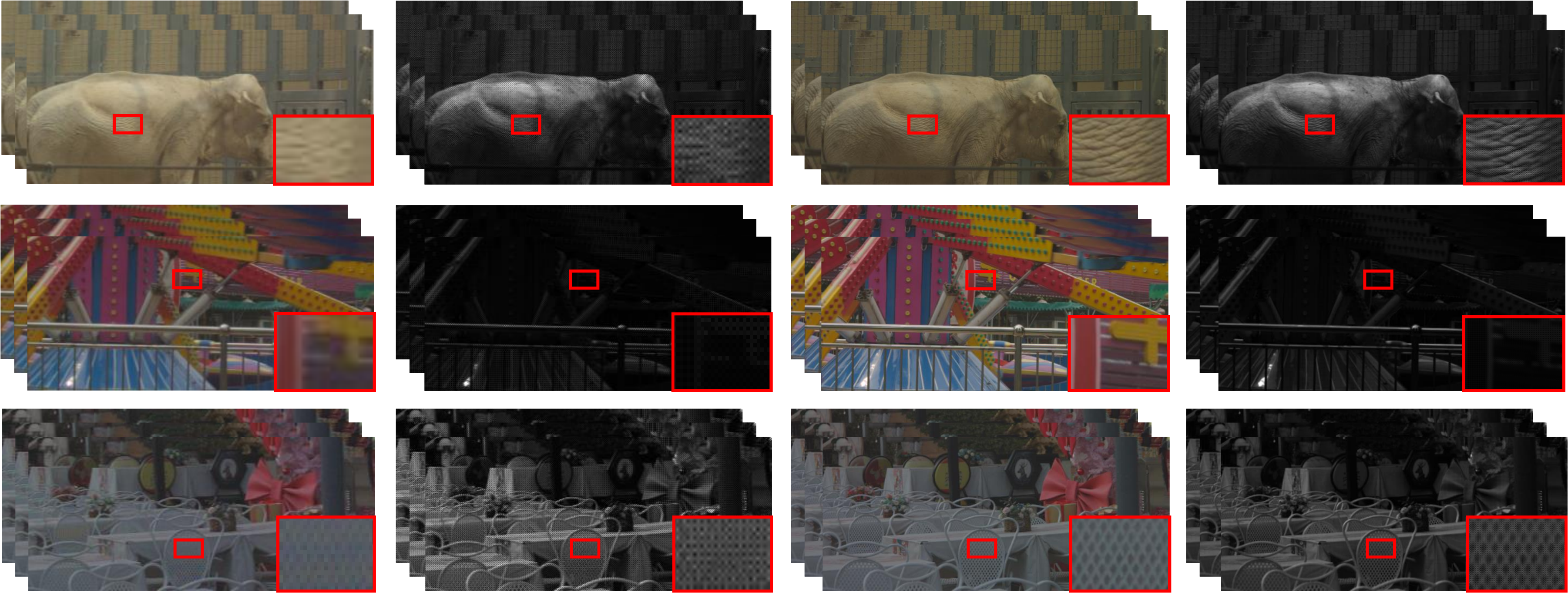}
   \caption{Examples of videos in Real-RawVSR Dataset with the brightness and contrast of raw frames adjusted for better visualization. From left to right, each column lists LR frames  ($I^{l_\text{rgb}}$, $I^{l_\text{raw}}$) and HR frames ($\tilde{I}^{h_\text{rgb}}$, $\tilde{I}^{h_\text{raw}}$) in both raw and sRGB domains.}
   \label{fig:fig2}
%   \vspace{-0.4cm}
\end{figure}
%------------------------------------------------------------------------
{\bf Data Processing.} 
As shown in Fig.~\ref{fig:fig1} (b), although there is no parallax between the LR-HR pair, the field of view (FoV) of the LR frame is much larger than that of the HR frame. In addition, due to the existence of lens distortion, there is still misalignment between the LR-HR pairs. Therefore, we utilize a coarse to fine alignment strategy to obtain aligned LR-HR pairs. In the following, we give details for sRGB frame and raw frame alignment, respectively.

1) \textbf{RGB frame alignment.} First, we estimate a homography matrix ($H$) between the upsampled LR ($\hat{I}^{l_{\text{rgb}}}$, the upsampling factor is estimated according to the ratio between the LR and HR focal lengths) and HR ($I^{h_{\text{rgb}}}$) frames using their matched SIFT \cite{lowe1999object} key points, which are selected by the RANSAC algorithm \cite{fischler1981random}. Note that, we perform alignment on $I^{h_{{\text{rgb}}}}$, to make the LR input of our network to be consistent with real captured LR frames, instead of performing alignment on $I^{l_{\text{rgb}}}$ as that in~\cite{yang2021real}. Then, the aligned HR frame is obtained by $\hat{I}^{h_{\text{rgb}}} = HI^{h_{\text{rgb}}}$. In this way, we can roughly crop the corresponding regions in the LR frame matched with the HR frame. Then, we utilize DeepFlow \cite{weinzaepfel2013deepflow}, which is a traditional flow estimation method, to perform pixel-wise alignment for the matching area. Finally, we crop the center area to eliminate the alignment artifacts around the border, generating the aligned LR-HR frames in RGB domain, denoted by ($I^{l_{\text{rgb}}}$, $\tilde{I}^{h_{\text{rgb}}}$).

2) \textbf{Raw frame alignment.} The raw frames should go through the same pipeline as that of RGB frames to make $\tilde{I}_t^{h_{\text{raw}}}$ and $\tilde{I}_t^{h_{\text{rgb}}}$  be strictly aligned. However, directly applying the global and local alignment will destroy the Bayer pattern of raw inputs. Therefore, we first pack the Bayer pattern raw frame into RGGB sub-frames, whose size is half of that of RGB frames. Hence, we change the $H$ matrix calculated from sRGB frames by rescaling the translation parameters with a ratio of 0.5. The deep flow vectors are also processed in the same way. In this way, we generate the raw frame pair ($I^{l_{\text{raw}}}$, $\tilde{I}^{h_{\text{raw}}}$). Note that, in this work, we utilize ($I^{l_{\text{raw}}}$, $\tilde{I}^{h_{\text{rgb}}}$) as training pairs. The provided raw pairs can enable future research on raw to raw SR.    

We totally captured 600 groups of videos, and manually removed 150 videos with large alignment errors, with 450 videos remaining in our dataset. Fig.~\ref{fig:fig2} gives some examples of our aligned pairs in both raw and sRGB domains. Note that, although they are aligned in spatial, there are still color and illumination differences in each LR-HR pair. These phenomena also exist in other real captured LR-HR pairs \cite{bhat2021deep,yang2021real,zhang2019zoom}. 
Our captured scenes vary from indoor to outdoor, and the motion types include camera motions and object motions. The resolution of the original HR frame is $1728\times972$. After alignment and center cropping, the resolutions of the aligned HR and LR frames for $2\times$ SR are $1440\times640$ and $720\times320$, respectively. For each magnification scale, there are 150 video pairs and each video contains about 150 frames. 
More detailed information about the dataset is presented in the supplementary file.

\section{The Proposed Method}
%------------------------------------------------------------------------
\begin{figure}[t]
  \centering
  \setlength{\abovecaptionskip}{0.cm}
   \includegraphics[width=1\textwidth]{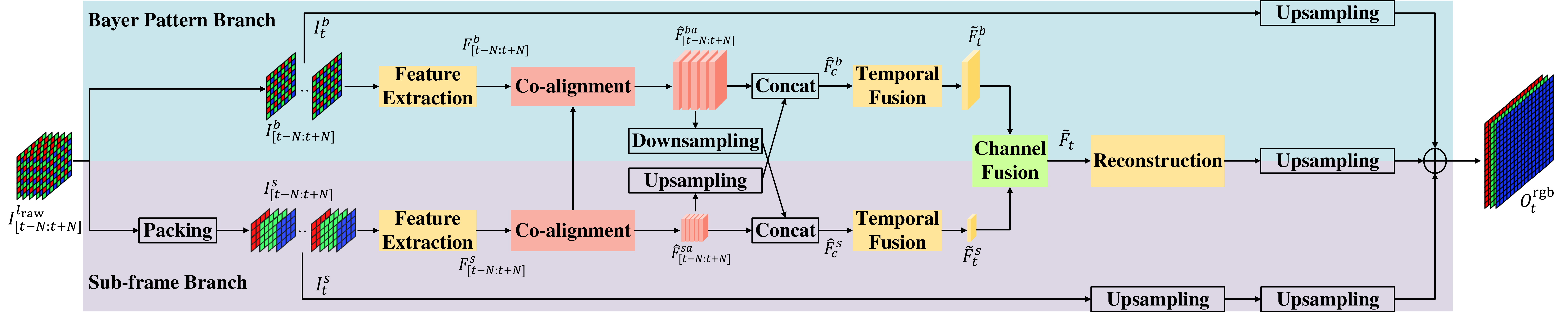}
   \caption{The proposed Real-RawVSR network for $2\times$ SR. The LR raw sequences ($I_{[t-N:t+N]}^{l_{\text{raw}}}$) are fed into the network in terms of both Bayer pattern and sub-frame forms. The final SR result $O_t^{\text{rgb}}$ is obtained by feature interaction and fusion of the Bayer pattern and sub-frame branches.}
   \label{fig:fig3}
%   \vspace{-0.4cm}
\end{figure}
%------------------------------------------------------------------------
We propose a Real-RawVSR network to reconstruct an HR sRGB frame $O_t^{\text{rgb}}$ from $2N+1$ consecutive LR raw frames $I_{[t-N:t+N]}^{l_{\text{raw}}}$. The existing raw image (video) SR methods~\cite{liu2021exploit,xu2019towards} usually directly pack the Bayer pattern input into four (RGGB) different channels, where each channel contains the same color pixels. However, this will destroy the pixel order of the original raw frame. Inspired by \cite{liang2020raw}, in this work, we propose to deal with raw frames in two branches, as shown in Fig.~\ref{fig:fig3}. The top branch deals with the original Bayer pattern input, and the bottom branch deals with the packed RGGB input.
In this way, the top Bayer pattern branch benefits the spatial reconstruction while the bottom sub-frame branch can take advantage of longer neighboring pixels to generate details. 
To fully take advantage of the complementary information between the two branches, we propose co-alignment, interaction, and fusion modules. In the following, we give details of these modules.

%The features of the two branches go through co-alignment, interaction, temporal fusion, channel fusion, reconstruction and upsampling modules to generate the HR sRGB outputs. 

%Inspired by the \cite{liang2020raw}, we input the unpacked single-channel bayer pattern $I^{b}_{[t-2:t+2]}$ and the packed RGGB $I^{s}_{[t-2:t+2]}$ into the network, as shown in Fig.~\ref{fig:fig3}. Note that, $I^{b}_{[t-2:t+2]}$ with size $H \times W$ is twice $I^{s}_{[t-2:t+2]}$. Inspired by the DCN module~\cite{dai2017deformable}, we propose the guided alignment module which develop the PCD alignment for the two raw inputs and a bidirectional interactive module enhancing the ability of the two branches. Then, the two branch features after guided alignment and information interaction pass through the channel fusion module, the reconstruction module, the upsampling module with a long skip-connection to obtain the restored HR sRGB output. The details are clarified in the following sections.
\subsection{Packing and Feature Extraction}
As shown in Fig.~\ref{fig:fig3}, the input LR raw frames $I_{[t-N:t+N]}^{l_{\text{raw}}}$ are fed into the network in different forms for the two branches. The top Bayer pattern branch directly utilizes the raw frames themselves as input. The bottom sub-frame branch utilizes the packed version, namely that we extract the sub-frame with the same color from the Bayer pattern input and all the sub-frames form a new sequence. For simplicity, we denote the input of the Bayer pattern branch as 
$I^b_{[t-N:t+N]}$ and that of the sub-frame branch as  $I^{s}_{[t-N:t+N]}$, whose channel number is four times of $I^b_{[t-N:t+N]}$. The Bayer pattern branch keeps the original order of raw pixels, which is good for spatial reconstruction. Although the sub-frame branch cannot keep the original pixel order, it can take advantage of far neighbor correlations to generate details. Therefore, they are complementary to each other, which helps to improve the SR results generated by one single branch. Then, the two inputs go through the feature extraction modules, respectively, where the feature extraction module is constructed by five residual blocks. Note that, the weights for the two feature extraction blocks are not shared since their inputs are in different forms.  
%Most networks consider that the output is a sRGB image and adopt the four-channel RGGB input after the pack so that the network can obtain stronger color information from the raw image, but this will destroy the original spatial structure of the raw data. Essentially, each pixel value of raw data is directly obtained by the camera sensor through a filter and photoelectric conversion, and there is no intensity correlation between each pixel. 
%Using the single-channel raw data as input can learn stronger spatial information and intensity information of neighboring pixels. Based on the above considerations, in order to be able to retain the spatial information of the raw data while learning more RGB color information, we input the packed raw data and  the single-channel raw data into the network. The two branches can extract richer raw information from two different angles of space and color and carry out information interaction so that the network can mine the deep information of raw as much as possible.
After the feature extraction module, we obtain $F_{[t-N:t+N]}^{b}$ with size $(2N+1)\times C\times H\times W$ for the Bayer pattern branch and $F_{[t-N:t+N]}^{s}$ with size $ (2N+1)\times C\times H/2\times W/2$ for the sub-frame branch, where $2N+1$ is the frame number along the time dimension, $C$ is the channel number, $H$ is the height, and $W$ is the width of features. 
%------------------------------------------------------------------------
\begin{figure}[t]
  \centering
  \setlength{\abovecaptionskip}{0.cm}
   \includegraphics[height = 6cm]{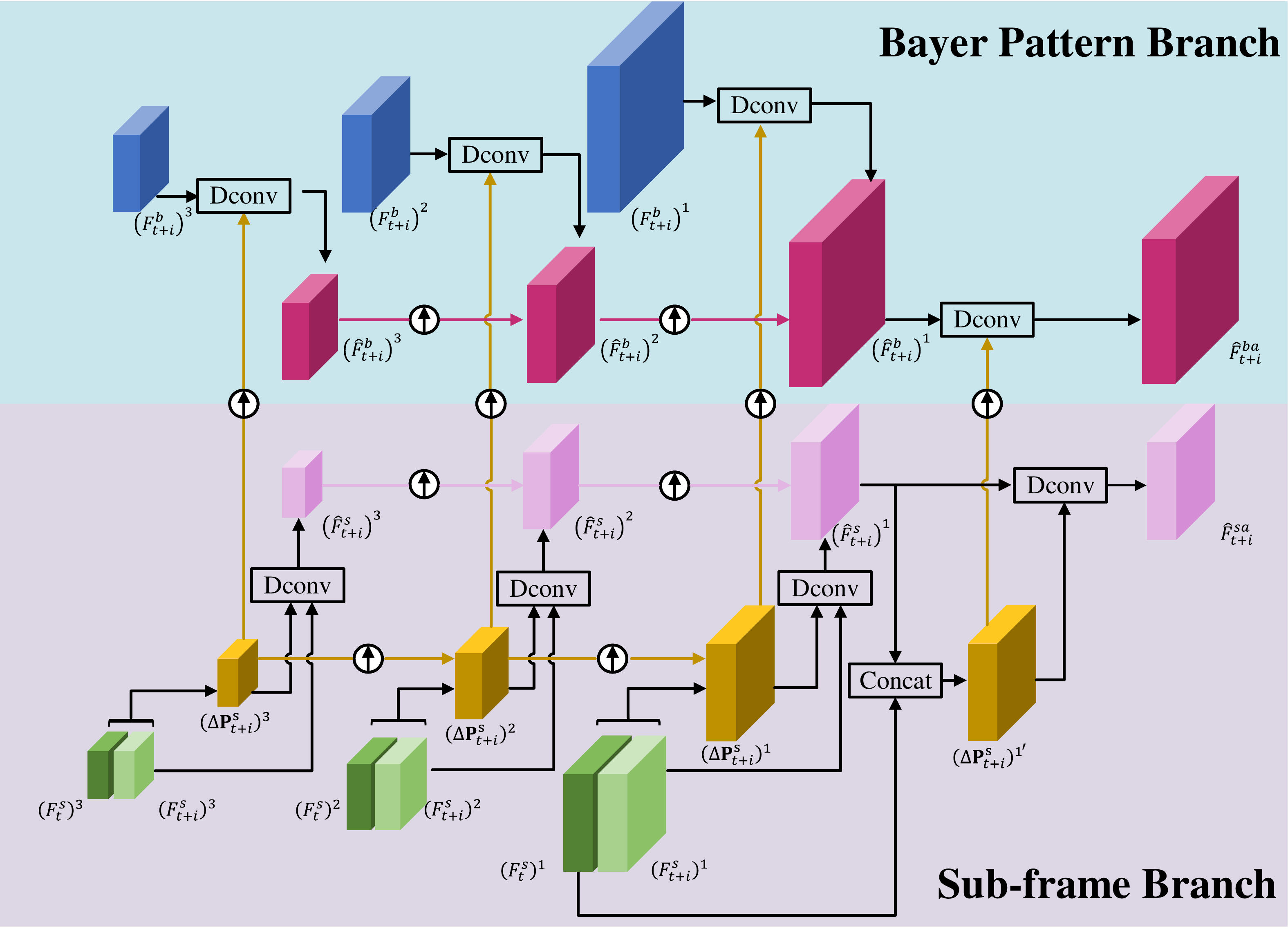}
   \caption{The proposed co-alignment module. The top branch is for the Bayer pattern feature alignment and the bottom branch is for the sub-frame feature alignment. The two branches share the same offset with different sizes.}
   \label{fig:fig4}
%   \vspace{-0.3cm}
\end{figure}
%------------------------------------------------------------------------
\subsection{Co-Alignment}
Since there are temporal misalignments between neighboring frames, we need to warp neighboring frames to the center frame. Following \cite{wang2019edvr}, we utilize PCD alignment. Since we have two branches, a straightforward solution is performing the PCD alignment separately. We note that the two branches actually share the same offset. Therefore, we propose to calculate the alignment offsets from the sub-frame branch and then directly copy the calculated offsets to the Bayer pattern branch to perform the alignment operation. Namely that the two branches are co-aligned. 

%We use the deformable convolution~\cite{dai2017deformable} to achieve the two-branch alignment. Specifically, we first perform PCD alignment~\cite{wang2019edvr} in the bottom pack branch. Then, we implement the $2\times$ upsampling at the offsets calculated from the three level pyramids and apply them to the top unpack branch. With the proposed alignment method, we can simultaneously align the features of the top and bottom branches from coarse to fine. This design can get the more accurate offset estimation and make full use of the feature relationship between the top and bottom branch to reduce the computation of alignment.

Given the features of two adjacent frames in the sub-frame branch $F_{t}^{s}$ and $F_{t+i}^{s}$, we aim to align $F_{t+i}^{s}$ with $F_{t}^{s}$. The aligned feature $\hat{F}_{t+i}^{s}$ at position $\textbf{p}_{0}$ is obtained via deformable convolution, which can be expressed by
\begin{equation}
    \begin{aligned}
        \hat{F}_{t+i}^{s}(\textbf{p}_{0}) = \sum_{k=0}^K w_{k} \cdot F_{t+i}^{s}(\textbf{p}_{0} + \textbf{p}_{k} + \triangle{\textbf{p}}_{k}) \cdot \triangle{m}_{k},
    \end{aligned}  
    \label{eq:1}
\end{equation}
where $\textbf{w}_{k}$ and $\textbf{p}_{k}$ represent the weight and predefined offset for the $k$-th location in the deformable convolution kernel. The learnable offset $\triangle{\textbf{p}}_{k}$ and the modulation scalar $\triangle{m}_{k}$ are predicted from concatenated features of the neighboring and reference frames, denoted by 
%------------------------------------------------------------------------
\begin{equation}
    \begin{aligned}
        {\triangle{\textbf{P}}}_{t+i} = f([F_{t+i}^{s}, F_{t}^{s}]),
    \end{aligned}  
    \label{eq:2}
\end{equation}
%------------------------------------------------------------------------
where $\triangle{\textbf{P}}=\left\{\triangle{\textbf{p}}\right\}$ represents the set of offsets, and $f$ represents a nonlinear mapping function realized by several convolution layers. 
For simplicity, we ignore the modulation scalar $\triangle{m}_{k}$ in the descriptions and figures.
Following PCD alignment, we further utilize pyramidal processing and cascading refinement to deal with large motions, as shown in Fig.~\ref{fig:fig4}. The features $(F_{t+i}^{s})^{l}$ and $(F_{t}^{s})^{l}$ are downsampled via strided convolution for $L-1$ times to form a pyramid with $L$ levels. The pyramid features in the Bayer-pattern branch are constructed in the same way. 
The offsets in the $l^{\text{th}}$ level are calculated from the concatenated features in the $l^{\text{th}}$ level and the upsampled version of the offsets in the $(l+1)^{\text{th}}$ level. The upsampling is realized by bilinear interpolation and the offset values are magnified by two times. 
%The top part implement the pack branch alignment, and the bottom part use the offsets upsampled from the top part to implement the unpack branch alignment. $(F_{t}^{s})^{l}$, $(F_{t+i}^{s})^{l}$ and $(F_{t+i}^{b})^{l}$ represents the $l$-level pack center feature, pack adjacent feature and unpack adjacent feature in the pyramid, respectively. In the top part, the offsets $(\triangle{\textbf{P}}_{t+i}^{s})^{l}$ in the $l$-level are calculated from both the upsampled offsets from the $(l+1)$-level and the $l$-level features. 
This process is denoted by
%------------------------------------------------------------------------
\begin{equation}
    \begin{aligned}
        (\triangle{\textbf{P}}_{t+i}^{s})^{l} = f([(F_{t+i}^{s})^{l}, (F_{t}^{s})^{l}], 2((\triangle{\textbf{P}}_{t+i}^{s})^{l+1})^{\uparrow2}).
    \end{aligned}  
    \label{eq:3}
\end{equation}

%------------------------------------------------------------------------
Since the input of the sub-frame branch is actually a down-sampling version of that in the Bayer pattern branch, the offset values for the Bayer pattern branch should be two times of that in the sub-frame branch. Therefore, the offsets for the 
Bayer pattern branch  $(\triangle{\textbf{P}}_{t+i}^{b})^{l}$ in the $l^{\text{th}}$ level can be obtained via two times upsampling and two times magnification of 
the offsets $(\triangle{\textbf{P}}_{t+i}^{s})^{l}$ in the sub-frame branch. We denote this process as 
%------------------------------------------------------------------------
\begin{equation}
    \begin{aligned}
        (\triangle{\textbf{P}}_{t+i}^{b})^{l} = 2((\triangle{\textbf{P}}_{t+i}^{s})^{l})^{\uparrow2}.
    \end{aligned}  
    \label{eq:4}
\end{equation}
%------------------------------------------------------------------------
Given the offsets, the aligned features for the two branches can be expressed by
%------------------------------------------------------------------------
\begin{equation}
    \begin{aligned}
        (\hat{F}_{t+i}^{s})^{l} = g(\text{Dconv}((F_{t+i}^{s})^{l}, (\triangle{\textbf{P}}_{t+i}^{s})^{l}), ((\hat{F}_{t+i}^{s})^{l+1})^{\uparrow2}),
    \end{aligned}  
    \label{eq:5}
\end{equation}
%------------------------------------------------------------------------
\begin{equation}
    \begin{aligned}
        (\hat{F}_{t+i}^{b})^{l} = g(\text{Dconv}((F_{t+i}^{b})^{l}, (\triangle{\textbf{P}}_{t+i}^{b})^{l}), ((\hat{F}_{t+i}^{b})^{l+1})^{\uparrow2}),
    \end{aligned}  
    \label{eq:6}
\end{equation}
%------------------------------------------------------------------------
where $g$ represents the mapping function realized by seveal convolution layers and $\text{DConv}$ represents deformable convolution expressed in Eq. \ref{eq:1}. Note that, the two-branch $\text{DConv}$ shares the same weights in the corresponding level. After alignment for $L$ levels, we further use the offsets $(\triangle{\textbf{P}}_{t+i}^{s})^{1'}$ calculated between $(F_{t}^{s})^1$ and $(\hat{F}_{t+i}^{s})^1$ to refine $(\hat{F}_{t+i}^{s})^1$ and $(\hat{F}_{t+i}^{b})^1$, and generate the final alignment results $\hat{F}_{t+i}^{sa}$ and $\hat{F}_{t+i}^{ba}$ for the neighboring features in the two branches. 

We would like to point out that using the proposed co-alignment strategy not only reduces computing complexity but also improves the final SR performance (see the ablation study). The main reason is that the offsets are optimized by both the Bayer pattern features and the sub-frame features, while the offsets calculated with separated alignment can only be optimized with their corresponding features. Therefore, the co-alignment strategy outperforms the sep-alignment.   
%------------------------------------------------------------------------
\subsection{Interaction}
Since the features in the two branches are complementary, we further propose an interaction module to enrich the feature representations in the two branches. Specifically, the Bayer pattern branch features are downsampled via a $3\times3$ strided convolution (stride=2) and Leaky Relu layer, and these downsampled features are concatenated with those in the sub-frame branch. Similarly, the sub-frame branch features are upsampled via pixel shuffle \cite{shi2016real}, which are then concatenated with the features in the Bayer pattern branch. In this way, we generate the interacted features 
$\hat{F}_{c}^{b}\in \mathbb{R}^{(4N+2)\times C\times H\times W}$ and $\hat{F}_{c}^{s} \in \mathbb{R}^{(4N+2)\times C\times H/2 \times W/2}$.

%add a bidirectional interactive module between the two branches. Due to the different sizes of the top and bottom branches, we concatenate the pack features with the unpack features processed by a $3\times3$ conv downsampling, while concatenating the unpack features with the pack features processed by a upsampling operation. Then, we obtain $\hat{F}_{\text{concat}}^{b}$ with size $B \times 2N \times C \times H \times W$ and $\hat{F}_{\text{concat}}^{s}$ with size $B \times 2N \times C \times \frac{H}{2} \times \frac{W}{2}$.
%------------------------------------------------------------------------
\subsection{Temporal Fusion}
Although we have aligned the neighboring frames to the reference frame, these frames still contribute differently to the reference frame SR. 
Therefore, we utilize attention based fusion to fuse the features together. First, we utilize a non-local temporal attention module \cite{yue2020supervised} to aggregate long-range features to enhance the feature representations along the time dimension. Then, we utilize temporal spatial attention (TSA) \cite{wang2019edvr} based fusion to fuse the features together. Finally, we obtain the temporal fused features $\tilde{F}_{t}^{b}$ with size $1 \times C \times H \times W$ and $\tilde{F}_{t}^{s}$ with size $1 \times C \times H/2 \times W/2 $ for the two branches, respectively. 

%We perform temporal attention and fusion on $2N$ temporal features in the top and bottom branches respectively to obtain the intermediate frame feature. Specifically, we utilize the strategy from~\cite{yue2020supervised} to process tempotal attention. For $2N$ features in the time dimension, we use the softmax function to calculate the weight between one and other features and update the information of the current feature by weighted summation.

%Then we adopt the TSA Fusion proposed in~\cite{wang2019edvr} to adaptively fuse $2N$ features. Note that, among the $2N$ features of each branch, we take the middle feature of the original N features in the current branch as the reference feature to execute the $2N$ TSA Fusion. Finally, we obtain the features: $\tilde{F}_{t}^{b}$ with size $B \times C \times H \times W$ and $\tilde{F}_{t}^{s}$ with size $B \times C \times \frac{H}{2} \times \frac{W}{2}$.
%------------------------------------------------------------------------
%------------------------------------------------------------------------
%\begin{figure}[t]
%  \centering
%   \includegraphics[width=0.8\textwidth]{figure/fig5.jpg}
%   \caption{Channel Fusion Module}
%   \label{fig:fig5}
%\end{figure}
%------------------------------------------------------------------------
\subsection{Channel Fusion}
We utilize channel fusion to merge the features in the two branches together since the same channel of  $\tilde{F}_{t}^{b}$ and $\tilde{F}_{t}^{s}$ may contribute differently to the final SR reconstruction. We adopt selective kernel convolution (SKF) \cite{li2019selective} to fuse the two branches via channel-wise weighted average. We first upsample  $\tilde{F}_{t}^{s}$ via pixel shuffle to make it have the same size as that of $\tilde{F}_{t}^{b}$. Then, the two features are added together, going through global average pooling along the channel dimension, generating a channel-wise weighting vector $z\in \mathbb{R}^{1\times1\times C}$. Then, $z$ goes through the squeeze and excitation layers, generating two weighting coefficients $z^b$ and $z^s$. Hereafter, they are normalized via softmax, generating the final weighting coefficients $\hat{z}^b$ and $\hat{z}^s$. The final fused feature is obtained by $\tilde{F}_t = \hat{z}^{b}\tilde{F}_{t}^{b} +\hat{z}^{s}\tilde{F}_{t}^{s}$.

%process channel fusion that specifically handles the features of the two branches. The relationship between the input size of the top and bottom branches is reflected in the fact that the receptive field is different during the convolution operation. Selective kernel convolution can adapt to the different receptive field and automatically select the effective feature channels in the two branches. This design can effectively select the effective channel information of the two features to achieve better information fusion.

%As shown in Fig.~\ref{fig:fig5}, We first execute the element-wise addition between $\tilde{F}_{t}^{b}$ and $\tilde{F}_{t}^{s}$. Note that, $\tilde{F}_{t}^{s}$ doubles the size in advance through the pixelshuffle module~\cite{shi2016real}. Then after the global average pooling which outputs the channel-wise vector $z$ with size $B \times C$, some simple fully connected layers and the softmax operation are used to obtain two attention coefficents $z^{b}$ and $z^{s}$. Finally, we get the fused feature $\tilde{F}_t$ through the processing denoted by:
%------------------------------------------------------------------------
%\begin{equation}
%    \begin{aligned}
%        \tilde{F}_t = z^{b} \cdot \tilde{F}_{t}^{b} + z^{s} \cdot \tilde{F}_{t}^{s} 
%    \end{aligned}  
 %   \label{eq:7}
%\end{equation}
%------------------------------------------------------------------------
\subsection{Reconstruction and Upsampling}
The fused feature $\tilde{F}_t$ is fed into the reconstruction module, which is realized by 10 ResNet blocks, for the SR reconstruction. After reconstruction, we utilize the pixel shuffle layer to upsample it and then utilize a convolution layer to generate the three-channel output. We also utilize two long skip connections.
One is for the LR Bayer input ($I_t^b$), which is first processed by a convolution layer and then upsampled by pixel shuffle to a three channel output. The other is for the LR sub-frame input ($I_t^s$), which is upsampled two times since its spatial size is half of the original input. 
The three outputs are added together to generate the final HR result $O_t^{\text{rgb}}$. For $4\times$ magnification, similar to EBSR~\cite{luo2021ebsr}, we utilize a two-stage upsampling based long-skip connection.   

%After extracting deep features from LR sequences, we use 10 resnet blocks to implement the reconstruction module. For $4\times$ VSR, we use two pixelshuffle modules to process the upsampling operation. In particular, motivated by EBSR~\cite{luo2021ebsr}, we employ a two-stage upsampling long-skip connection between the lr features and the network outputs, and add the two-stage upsampling results to the upsampling features in the main network. This design is more conducive to the main network to learn the detail residuals and achieve better performance. Finally, we get the center frame VSR result: $\hat{O}_t^{\text{rgb}}$ with size $3 \times 4H \times 4W$.
\subsection{Color Correction and Loss Function}
\label{sec:loss}
As described in Sec.~\ref{sec:dataset}, the LR input ($I_t^{l_{\text{rgb}}}$) and ground truth ($\tilde{I}_t^{h_{\text{rgb}}}$) have differences in color and brightness. Directly utilizing pixel-wise loss between the output and the ground truth may lead the network to optimize color and brightness correction other than the essential task of SR, \textit{i.e.,} detail generation. To solve this problem, inspired by \cite{bhat2021deep}, we utilize color correction before the loss calculation. Different from \cite{bhat2021deep}, we utilize channel-based color correction for RGB channels separately other than calculating a $3\times 3$ color correction matrix to simultaneously correct them. This process can be denoted as 
\begin{equation}
{\hat{O}}_t^c = \alpha^c O_t^c, \alpha^c = \phi(I_t^{l_c}, {\tilde{I}}_t^{h_c}), c\in\{r,g,b\},
\label{eq:7}
\end{equation}
where $\alpha^c$ is the scaling factor for channel $c$, and it is calculated by minimizing the least square loss between the corresponding pixel pairs in $I_t^{l_c}$ and the downsampled version of ${\tilde{I}}_t^{h_c}$. 
\iffalse
Since we use two cameras to capture the data, there are obvious differences in color and brightness between the data pairs. The pixel-wise loss is particularly sensitive to the color and intensity of the pixel. In order to solve this problem, we adopted a novel color conversion method. Unlike DBSR, we calculate the color conversion coefficient for the three color channels RGB between $I^{\text{rgb}}_t$ and $GT^{\text{rgb}}_t$ by by minimizing a least squares loss and use them in the three output channels of RGB, respectively. This conversion process can be expressed as:
%------------------------------------------------------------------------
\begin{equation}
    \begin{aligned}
        &{\hat{O}}_t^r = C^r(I_t^r, {GT}_t^r) \cdot O_t^r\\
        &{\hat{O}}_t^g = C^g(I_t^g, {GT}_t^g) \cdot O_t^g\\
        &{\hat{O}}_t^b = C^b(I_t^b, {GT}_t^b) \cdot O_t^b
    \end{aligned}  
    \label{eq:8}
\end{equation}
%------------------------------------------------------------------------
where $C^r$, $C^g$, $C^b$ represent the R, G, B coefficient, respectively.
\fi
Then, we can optimize the network with the Charbonnier loss~\cite{lai2017deep} between the corrected output and the ground truth as $\mathcal{L}= \sqrt{\parallel{\hat{O}}^{\text{rgb}}_t - {\tilde{I}}_t^{h_\text{rgb}}\parallel^2_2 + \epsilon}$, where $\epsilon=1 \times 10^{-6}$. 
\iffalse
 %${\hat{O}}^{\text{rgb}}_t$ and the ground truth ${\tilde{I}}_t^{h_\text{rgb}}$ 
%------------------------------------------------------------------------
\begin{equation}
    \begin{aligned}
        \mathcal{L}=\frac{1}{N}\sum_{t=0}^N \sqrt{\parallel{\hat{O}}^{\text{rgb}}_t - {\tilde{I}}_t^{h_\text{rgb}}\parallel^2_2 + \epsilon} 
    \end{aligned}
    \label{eq:9}
\end{equation}
%------------------------------------------------------------------------
where $\epsilon=1 \times 10^{-6}$, and N denotes the batch size.
\fi
%------------------------------------------------------------------------

%------------------------------------------------------------------------
\section{Experiments}
\subsection{Training Details}
In our experiments, for each magnification factor, 130 videos are used for training and validation, and the other 20 videos are used for testing. To make the movements between neighboring frames more obvious, for each video, we extract frames from the original 150 frames with a step size of three, resulting in a 50-frame sequence. This strategy is also used in \cite{nah2019ntire}. The raw data is pre-processed by black level subtraction and white level normalization.
%All experiments are trained and tested on our proposed Real-RawVSR dataset. The dataset contains 450 videos, \textit{i.e.} 150 videos at each magnification. The 150 videos are split into 130 for training set and 20 for testing set. In order to avoid the repetition of adjacent frames in a single video, we extract frames at 3 equal intervals based on about 150 frames of the original video sequence, which means each video sequence contains about 50 frames.
The frame number is 5, \textit{i.e.,} $N=2$. The channel number $C$ of features is $64$. All the convolution filter size is $3\times3$ \footnote{More details about the network structure are presented in the supplementary file.}. During training, the Bayer pattern patch size is $128\times128$ and the batch size is 4. We train our model with Adam optimizer and the learning rate is set to 1e-4. The total iteration number is 300k. Our model is implemented in PyTorch and trained with an NVIDIA 3090 GPU.
%------------------------------------------------------------------------

%------------------------------------------------------------------------
\subsection{Comparison with State-of-the-arts}
We compare with six state-of-the-art VSR methods, including four methods in sRGB domain (TOFlow~\cite{xue2019video}, EDVR~\cite{wang2019edvr}, TDAN~\cite{tian2020tdan}, and BasicVSR~\cite{chan2021basicvsr}) and two methods in raw domain (RawVSR \cite{liu2021exploit} and DBSR~\cite{bhat2021deep}). In addition, we also revise EDVR by setting its input to the one channel Bayer pattern input and the original bilinear upsampling operation on the long skip connection is replaced by convolution and pixel shuffle operations. The revised version is denoted as RawEDVR. For a fair comparison, we retrain the above methods on our dataset and add the color correction strategy mentioned in Sec.~\ref{sec:loss} to all the compared methods to avoid the influence of color mis-matching. We use $(I_{[t-N:t+N]}^{l_{\text{rgb}}}, \tilde{I}_t^{h_{\text{rgb}}})$ as training pairs for sRGB domain methods and $(I_{[t-N:t+N]}^{l_{\text{raw}}}, \tilde{I}_t^{h_{\text{rgb}}})$ for raw domain methods. All the methods are trained with 5 consecutive frames as inputs.  

%sRGB data as input for the sRGB models, and preprocessed raw data as input for the RawVSR and DBSR models. Note that, we need to change the upsampling module for different magnifications. During retraining the BasicVSR model, we use 5 frames for training in each iteration, and use the whole sequence in each video for testing following the strategy in~\cite{chan2021basicvsr}.

\begin{figure}[t]
    \centering
    \setlength{\abovecaptionskip}{0.cm}
    \includegraphics[width=0.85\textwidth]{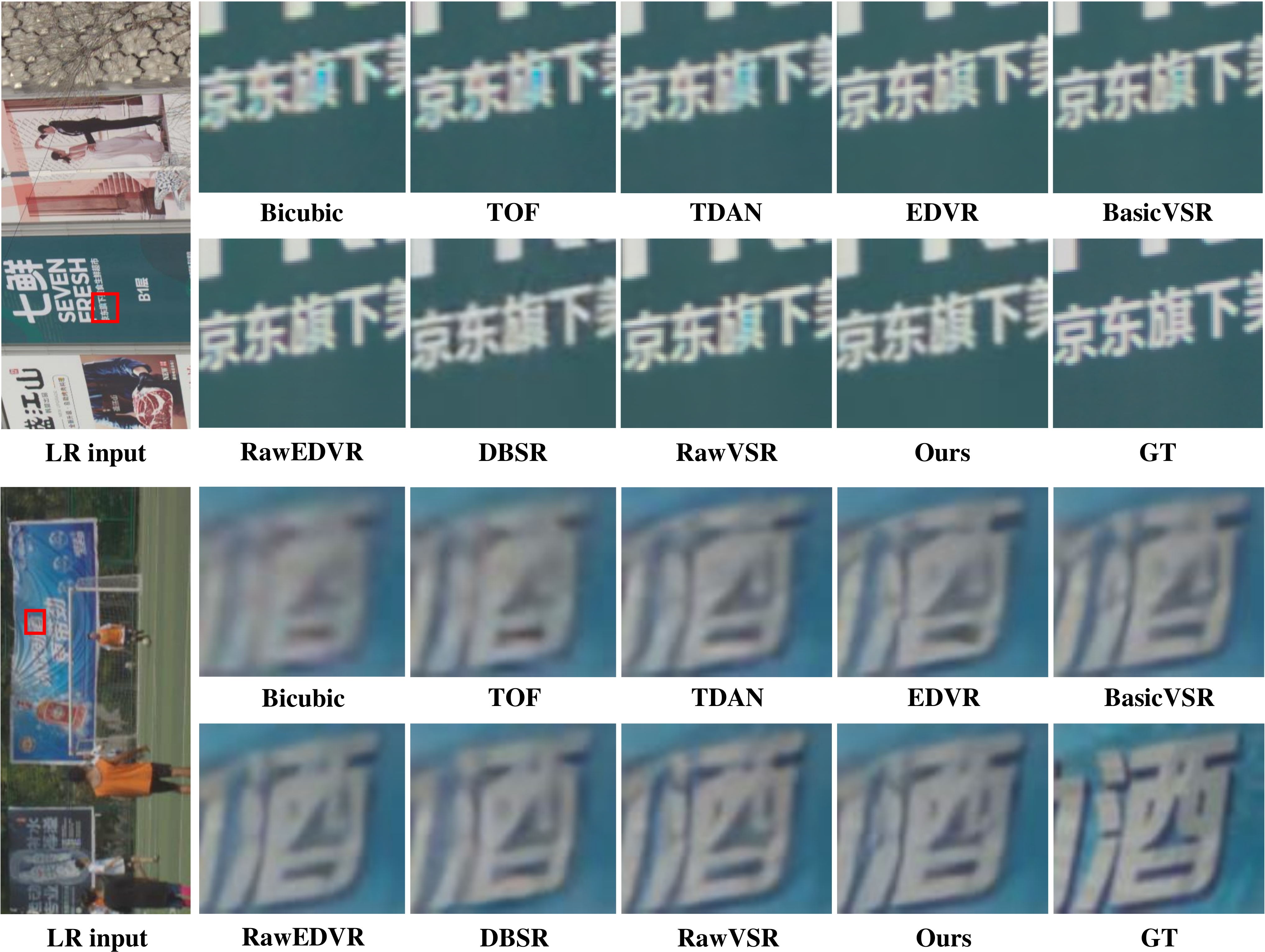}
    \caption{Visual comparison for $2\times$ and $4\times$ VSR results. For each group, the top (bottom) row presents the results generated by sRGB (raw) domain processing methods.}
    \label{fig:fig5}
  %  \vspace{-0.3cm}
\end{figure}

The quantitative comparison results are shown in Table~\ref{tab:tab1}. Our method achieves the best results compared to all previous methods on all scaling factors. Specially, for $2\times$ SR, our method outperforms EDVR and RawVSR by $0.45$ dB and $0.83$ dB, respectively. Note that, although the PSNR results of RawVSR and RawEDVR are worse than those of EDVR, the SSIM results of RawVSR and RawEDVR are generally better than those of EDVR. This demonstrates that the raw input is beneficial for the structure reconstruction, which is also verified by the visual comparison in Fig. \ref{fig:fig5}. We also present the number of parameters and FLOPs (calculated for $4\times$ SR with a $160\times360$ input) in Table~\ref{tab:tab1}. Our method has similar FLOPs as that of RawEDVR and is much lighter than RawVSR. This mainly benefits from the proposed co-alignment strategy, which saves about 100G FLOPs compared with separate alignment. 

\begin{table}[t]
	\centering
	\renewcommand\arraystretch{1.5}
	\caption{Quantitative comparison with state-of-the-art VSR methods. The best results are highlighted in bold and the second best results are underlined.}
    \resizebox{\textwidth}{!}{
    	\begin{tabular}{ccccccccccc}
    	    \toprule
    		\ Scale \ & &\ Bicubic\ &\ TOF~\cite{xue2019video}\ &\ TDAN~\cite{tian2020tdan}\ &\ EDVR~\cite{wang2019edvr}\ &\ BasicVSR~\cite{chan2021basicvsr}\ &\ RawEDVR\ &\ DBSR~\cite{bhat2021deep}\ &\ RawVSR~\cite{liu2021exploit}\ &\ Ours \ \\
    	    \midrule
            \multirow{2}{*}{\ $2\times$ \ } &\  PSNR \ & 35.32 & 35.62 & 36.14 & \underline{36.93} & 36.72 & 36.74 & 36.16 & 36.55 & \textBF{37.38} \\
            \multirow{2}{*}{} & SSIM & 0.9530 & 0.9555 & 0.9615 & 0.9674 & 0.9668 & 0.9670 & 0.9621 & \underline{0.9677} & \textBF{0.9705} \\
            \midrule
            \multirow{2}{*}{\ $3\times$ \ } & PSNR & 33.09 & 33.72 & 34.43 & \underline{35.25} & 34.95 & 35.23 & 34.48 & 34.96 & \textBF{35.62} \\
            \multirow{2}{*}{} & SSIM & 0.9169 & 0.9241 & 0.9352 & 0.9425 & 0.9408 & \underline{0.9442} & 0.9370 & 0.9431 & \textBF{0.9468} \\ 
            \midrule
            \multirow{2}{*}{\ $4\times$ \ } & PSNR & 31.19 & 32.17 & 32.84 & \underline{33.60} & 33.27 & 33.55 & 32.86 & 33.46 & \textBF{33.91} \\
            \multirow{2}{*}{} & SSIM & 0.8787 & 0.8928 & 0.9050 & 0.9139 & 0.9113 & 0.9153 & 0.9077 & \underline{0.9164} & \textBF{0.9182} \\
            \midrule
            \midrule
            \multicolumn{2}{c}{Params (M)} & - & - & 2.3 & 3.3 & 6.3 & 3.3 & 12.4 & 4.5 & 4.8 \\
            \multicolumn{2}{c}{FLOPs (G)} & - & - & 360.3 & 463.3 & 370.0 & 464.7 & 254.7 & 622.9 & 494.9 \\
            \bottomrule
    	\end{tabular}
    }
	\label{tab:tab1}
%	\vspace{-0.4cm}
\end{table}

%The main reason is that there is color and brightness differences between our LR-HR pair, which is hard to be learned for raw inputs. In contrast, our method is the best in terms of both PSNR and SSIM measurements.  

%It demonstrates that the method designed for synthetic LR raw sequences are not suitable for real LR raw sequences. One reason is that there is color and brightness differences between our LR-HR pair. Therefore, it is not suitable to utilize the LR sRGB frame to learn color correction module.  

%the PSNR gains of ours over the raw methods, DBSR and RawVSR, are 1.26dB and 0.69dB. Our method also surpasses the popular sRGB methods, BasicVSR and EDVR, by 0.84dB and 0.33dB. 

Fig.~\ref{fig:fig5} presents the visual comparison results for $2\times$ and $4\times$ SR. All the sRGB domain processing methods cannot deal with the false colors embedded in the LR input. RawVSR also cannot remove the false colors since it utilizes the LR sRGB input for guidance. It demonstrates that for real raw VSR, utilizing the LR sRGB input as guidance may be not a good choice. In addition, raw domain processing can generate better details compared with sRGB domain processing (see the second image). Our method can correct the false colors well and our generated details are much cleaner than those of other methods.  

%It can be observed that our method can correct the false colors and recover sharp details. 

%It demonstrates that the raw domain processing is beneficial for high frequency detail generation.  
%------------------------------------------------------------------------
\begin{table}[t]
	\centering
% 	\scriptsize
	\renewcommand\arraystretch{1.2}
	\caption{Ablation study ($4\times$) for the key modules in our network.}
	\resizebox{\textwidth}{!}{
    	\begin{tabular}{llccc}
    	    \toprule
    		\multirow{3}{*}{Alignment} & Sep-alignment & \ding{55} & \checkmark & \ding{55} \\
    		\multirow{3}{*}{} & Co-alignment & \ding{55} & \ding{55} & \checkmark \\
    		\multirow{3}{*}{} & PSNR/SSIM &\ 33.72/0.9173 \ &\  33.80/0.9178 \ &\  \textBF{33.91/0.9182} \ \\
    		\midrule
    		\multirow{2}{*}{Interaction} & Interaction & \ding{55} & \checkmark &  \\
    		\multirow{2}{*}{} & {PSNR/SSIM} & 33.76/0.9173 & \textBF{33.91/0.9182} &  \\
    		\midrule
    		\multirow{3}{*}{Channel Fusion} & SKF & \ding{55} & \checkmark & \\
    		\multirow{3}{*}{} & Concat & \checkmark & \ding{55} & \\
    		\multirow{3}{*}{} & PSNR/SSIM & 33.85/0.9180 & \textBF{33.91/0.9182} & \\
    		\midrule
    		\multirow{3}{*}{Color Correction} & Matrix-based & \ding{55} & \checkmark & \ding{55} \\
    		\multirow{3}{*}{} & Channel-based & \ding{55} & \ding{55} & \checkmark \\
    		\multirow{3}{*}{} & PSNR/SSIM & 30.65/0.9102 & 33.81/0.9173 & \textBF{33.91/0.9182} \\
    		\midrule
    		\multirow{3}{*}{Branch} & Bayer Branch & \checkmark & \ding{55} & \checkmark \\
    		\multirow{3}{*}{} &Sub-frame Branch & \ding{55} & \checkmark & \checkmark \\
    		%\multirow{4}{*}{} & PSNR/SSIM & 33.59/0.9161 & 33.46/0.9138 & \textBF{33.91/0.9182} \\
    		\multirow{3}{*}{} & PSNR/SSIM & 33.73/0.9175 & 33.70/0.9167 & \textBF{33.91/0.9182} \\
    		\bottomrule
    	\end{tabular}
    }
	\label{tab:tab2}
\end{table}
%------------------------------------------------------------------------
\subsection{Ablation Study}
% Table \ref{tab:tab2} lists the ablation study results for the key modules in our network. 
We evaluate the key modules in our network by replacing them with other straightforward solutions. 
1) \textbf{Co-Alignment.} We evaluate the effectiveness of the co-alignment module by replacing it with a separate alignment, which performs alignment on the two branches separately. As shown in Table~\ref{tab:tab2}, co-alignment outperforms sep-alignment by 0.11 dB. The main reason is that the offsets calculated by co-alignment are more accurate than those calculated by sep-alignment. We also present the result by removing the alignment module, which is 0.19 dB less than our proposed method. 2) \textbf{Interaction.} If we remove the interaction module from our full model, the result will drop 0.15 dB. It verifies that interaction is beneficial for taking advantage of the complementary information in the two branches. 3) \textbf{Channel Fusion.} By replacing the selective kernel based fusion strategy with simple concatenation, the PSNR result will drop 0.06 dB. 4) \textbf{Color Correction.} If we do not utilize color correction, the result will be heavily degraded (30.65 dB) due to the color cast. In addition, our channel-based correction is better than the widely used matrix-based correction method by 0.1 dB. 5) \textbf{Single Branch.} We also present the results by training the Bayer pattern branch and sub-frame branch separately. We increase the parameters of the two variants by increasing their channel numbers to make their parameters almost the same as that of our full model. Our method outperforms the two variants by nearly 0.2 dB. It demonstrates that the gain of two branch processing is not from the large parameters but from our co-alignment and interaction modules. 
%Visual comparison results for these ablation studies are in the supplementary file. 
%We train the top and bottom branches separately, retaining other modules except for interaction module and channel fusion. From Table.~\ref{tab:tab2}, we can see that the two-branch model has a gain of more than 0.3dB in PSNR values compared to using pack branch or unpack branch alone. This proves that the combination of pack and unpack branches can fully tap the potential information for raw data.

%We train the model using PCD, guided alignment module and without alignment module, respectively. Note that, while training the model using PCD, PCD alignment is used both pack and unpack branch. Table.~\ref{tab:tab2} shows guided alignment can get the best PSNR results. Specially, it is increased by 0.26 dB compared with the model without alignment.
%------------------------------------------------------------------------
\section{Conclusion and Discussion}
We build the first real-world raw VSR dataset with three magnification ratios in both raw and sRGB domains, which provides a benchmark dataset for both training and evaluation of real raw VSR methods. Based on this dataset, we propose a Real-RawVSR method by dealing with the raw inputs in two branches. By utilizing the proposed co-alignment, interaction, and fusion modules, the complementary information of the two branches is well explored. Experiments demonstrate that the proposed method outperforms state-of-the-art raw and sRGB VSR methods.   

Compared with VSR for synthetic LR inputs, dealing with real LR inputs is more difficult due to the color and brightness differences in the LR-HR pair. As reported in \cite{yang2021real}, the gap between different methods retrained on the same real dataset is much smaller than those trained on the synthetic dataset \cite{wang2019edvr}. In this work, we focus on the network structure design for raw inputs and have achieved impressive gain over our baseline network EDVR. The proposed co-alignment and interaction strategy can be applied to other sRGB VSR methods to improve their performance in dealing with raw inputs. In the future, we would like to explore more effective losses to deal with the color and brightness differences. 

%Note that, besides raw data in the Bayer pattern, our network can be extended to other CFA patterns. For X-Trans patterns, the raw frame can be packed in terms of $6\times6$ blocks, similar to that in \cite{chen2018learning}.

%Note that, the color and brightness differences may hinder the learning of high frequency details, even if we have utilized color correction before calculating the pixel-wise loss. In the future we would like to explore more effective losses for these real cases. 

\clearpage
% ---- Bibliography ----
%
% BibTeX users should specify bibliography style 'splncs04'.
% References will then be sorted and formatted in the correct style.
%
\bibliographystyle{splncs04}
\bibliography{egbib}

\begin{thebibliography}{10}
\providecommand{\url}[1]{\texttt{#1}}
\providecommand{\urlprefix}{URL }
\providecommand{\doi}[1]{https://doi.org/#1}

\bibitem{abdelhamed2020ntire}
Abdelhamed, A., Afifi, M., Timofte, R., Brown, M.S.: Ntire 2020 challenge on
  real image denoising: Dataset, methods and results. In: Proceedings of the
  IEEE/CVF Conference on Computer Vision and Pattern Recognition Workshops. pp.
  496--497 (2020)

\bibitem{abdelhamed2019ntire}
Abdelhamed, A., Timofte, R., Brown, M.S.: Ntire 2019 challenge on real image
  denoising: Methods and results. In: Proceedings of the IEEE/CVF Conference on
  Computer Vision and Pattern Recognition Workshops. pp.~0--0 (2019)

\bibitem{agustsson2017ntire}
Agustsson, E., Timofte, R.: Ntire 2017 challenge on single image
  super-resolution: Dataset and study. In: Proceedings of the IEEE conference
  on computer vision and pattern recognition workshops. pp. 126--135 (2017)

\bibitem{bhat2021deep}
Bhat, G., Danelljan, M., Van~Gool, L., Timofte, R.: Deep burst
  super-resolution. In: Proceedings of the IEEE/CVF Conference on Computer
  Vision and Pattern Recognition. pp. 9209--9218 (2021)

\bibitem{brooks2019unprocessing}
Brooks, T., Mildenhall, B., Xue, T., Chen, J., Sharlet, D., Barron, J.T.:
  Unprocessing images for learned raw denoising. In: Proceedings of the
  IEEE/CVF Conference on Computer Vision and Pattern Recognition. pp.
  11036--11045 (2019)

\bibitem{caballero2017real}
Caballero, J., Ledig, C., Aitken, A., Acosta, A., Totz, J., Wang, Z., Shi, W.:
  Real-time video super-resolution with spatio-temporal networks and motion
  compensation. In: Proceedings of the IEEE Conference on Computer Vision and
  Pattern Recognition. pp. 4778--4787 (2017)

\bibitem{cai2019ntire}
Cai, J., Gu, S., Timofte, R., Zhang, L.: Ntire 2019 challenge on real image
  super-resolution: Methods and results. In: Proceedings of the IEEE/CVF
  Conference on Computer Vision and Pattern Recognition Workshops. pp.~0--0
  (2019)

\bibitem{cai2019toward}
Cai, J., Zeng, H., Yong, H., Cao, Z., Zhang, L.: Toward real-world single image
  super-resolution: A new benchmark and a new model. In: Proceedings of the
  IEEE/CVF International Conference on Computer Vision. pp. 3086--3095 (2019)

\bibitem{chan2021basicvsr}
Chan, K.C., Wang, X., Yu, K., Dong, C., Loy, C.C.: Basicvsr: The search for
  essential components in video super-resolution and beyond. In: Proceedings of
  the IEEE/CVF Conference on Computer Vision and Pattern Recognition. pp.
  4947--4956 (2021)

\bibitem{chan2021basicvsr++}
Chan, K.C., Zhou, S., Xu, X., Loy, C.C.: Basicvsr++: Improving video
  super-resolution with enhanced propagation and alignment. arXiv preprint
  arXiv:2104.13371  (2021)

\bibitem{chan2021investigating}
Chan, K.C., Zhou, S., Xu, X., Loy, C.C.: Investigating tradeoffs in real-world
  video super-resolution. arXiv preprint arXiv:2111.12704  (2021)

\bibitem{chen2019camera}
Chen, C., Xiong, Z., Tian, X., Zha, Z.J., Wu, F.: Camera lens super-resolution.
  In: Proceedings of the IEEE/CVF Conference on Computer Vision and Pattern
  Recognition. pp. 1652--1660 (2019)

\bibitem{chen2019seeing}
Chen, C., Chen, Q., Do, M.N., Koltun, V.: Seeing motion in the dark. In:
  Proceedings of the IEEE/CVF International Conference on Computer Vision. pp.
  3185--3194 (2019)

\bibitem{chen2018learning}
Chen, C., Chen, Q., Xu, J., Koltun, V.: Learning to see in the dark. In:
  Proceedings of the IEEE Conference on Computer Vision and Pattern
  Recognition. pp. 3291--3300 (2018)

\bibitem{dai2017deformable}
Dai, J., Qi, H., Xiong, Y., Li, Y., Zhang, G., Hu, H., Wei, Y.: Deformable
  convolutional networks. In: Proceedings of the IEEE international conference
  on computer vision. pp. 764--773 (2017)

\bibitem{fischler1981random}
Fischler, M.A., Bolles, R.C.: Random sample consensus: a paradigm for model
  fitting with applications to image analysis and automated cartography.
  Communications of the ACM  \textbf{24}(6),  381--395 (1981)

\bibitem{jiang2019learning}
Jiang, H., Zheng, Y.: Learning to see moving objects in the dark. In:
  Proceedings of the IEEE/CVF International Conference on Computer Vision. pp.
  7324--7333 (2019)

\bibitem{joze2020imagepairs}
Joze, H.R.V., Zharkov, I., Powell, K., Ringler, C., Liang, L., Roulston, A.,
  Lutz, M., Pradeep, V.: Imagepairs: Realistic super resolution dataset via
  beam splitter camera rig. In: Proceedings of the IEEE/CVF Conference on
  Computer Vision and Pattern Recognition Workshops. pp. 518--519 (2020)

\bibitem{kappeler2016video}
Kappeler, A., Yoo, S., Dai, Q., Katsaggelos, A.K.: Video super-resolution with
  convolutional neural networks. IEEE transactions on computational imaging
  \textbf{2}(2),  109--122 (2016)

\bibitem{lai2017deep}
Lai, W.S., Huang, J.B., Ahuja, N., Yang, M.H.: Deep laplacian pyramid networks
  for fast and accurate super-resolution. In: Proceedings of the IEEE
  conference on computer vision and pattern recognition. pp. 624--632 (2017)

\bibitem{li2019selective}
Li, X., Wang, W., Hu, X., Yang, J.: Selective kernel networks. In: Proceedings
  of the IEEE/CVF Conference on Computer Vision and Pattern Recognition. pp.
  510--519 (2019)

\bibitem{liang2020raw}
Liang, C.H., Chen, Y.A., Liu, Y.C., Hsu, W.: Raw image deblurring. IEEE
  Transactions on Multimedia  (2020)

\bibitem{liu2019learning}
Liu, J., Wu, C.H., Wang, Y., Xu, Q., Zhou, Y., Huang, H., Wang, C., Cai, S.,
  Ding, Y., Fan, H., et~al.: Learning raw image denoising with bayer pattern
  unification and bayer preserving augmentation. In: Proceedings of the
  IEEE/CVF Conference on Computer Vision and Pattern Recognition Workshops.
  pp.~0--0 (2019)

\bibitem{liu2021exploit}
Liu, X., Shi, K., Wang, Z., Chen, J.: Exploit camera raw data for video
  super-resolution via hidden markov model inference. IEEE Transactions on
  Image Processing  \textbf{30},  2127--2140 (2021)

\bibitem{lowe1999object}
Lowe, D.G.: Object recognition from local scale-invariant features. In:
  Proceedings of the seventh IEEE international conference on computer vision.
  vol.~2, pp. 1150--1157. Ieee (1999)

\bibitem{lugmayr2020ntire}
Lugmayr, A., Danelljan, M., Timofte, R.: Ntire 2020 challenge on real-world
  image super-resolution: Methods and results. In: Proceedings of the IEEE/CVF
  Conference on Computer Vision and Pattern Recognition Workshops. pp. 494--495
  (2020)

\bibitem{luo2021ebsr}
Luo, Z., Yu, L., Mo, X., Li, Y., Jia, L., Fan, H., Sun, J., Liu, S.: Ebsr:
  Feature enhanced burst super-resolution with deformable alignment. In:
  Proceedings of the IEEE/CVF Conference on Computer Vision and Pattern
  Recognition. pp. 471--478 (2021)

\bibitem{nah2019ntire}
Nah, S., Baik, S., Hong, S., Moon, G., Son, S., Timofte, R., Mu~Lee, K.: Ntire
  2019 challenge on video deblurring and super-resolution: Dataset and study.
  In: Proceedings of the IEEE/CVF Conference on Computer Vision and Pattern
  Recognition Workshops. pp.~0--0 (2019)

\bibitem{niu2020single}
Niu, B., Wen, W., Ren, W., Zhang, X., Yang, L., Wang, S., Zhang, K., Cao, X.,
  Shen, H.: Single image super-resolution via a holistic attention network. In:
  European conference on computer vision. pp. 191--207. Springer (2020)

\bibitem{shi2016real}
Shi, W., Caballero, J., Husz{\'a}r, F., Totz, J., Aitken, A.P., Bishop, R.,
  Rueckert, D., Wang, Z.: Real-time single image and video super-resolution
  using an efficient sub-pixel convolutional neural network. In: Proceedings of
  the IEEE conference on computer vision and pattern recognition. pp.
  1874--1883 (2016)

\bibitem{tian2020tdan}
Tian, Y., Zhang, Y., Fu, Y., Xu, C.: Tdan: Temporally-deformable alignment
  network for video super-resolution. In: Proceedings of the IEEE/CVF
  Conference on Computer Vision and Pattern Recognition. pp. 3360--3369 (2020)

\bibitem{wang2019edvr}
Wang, X., Chan, K.C., Yu, K., Dong, C., Change~Loy, C.: Edvr: Video restoration
  with enhanced deformable convolutional networks. In: Proceedings of the
  IEEE/CVF Conference on Computer Vision and Pattern Recognition Workshops.
  pp.~0--0 (2019)

\bibitem{wang2020practical}
Wang, Y., Huang, H., Xu, Q., Liu, J., Liu, Y., Wang, J.: Practical deep raw
  image denoising on mobile devices. In: European Conference on Computer
  Vision. pp. 1--16. Springer (2020)

\bibitem{weinzaepfel2013deepflow}
Weinzaepfel, P., Revaud, J., Harchaoui, Z., Schmid, C.: Deepflow: Large
  displacement optical flow with deep matching. In: Proceedings of the IEEE
  international conference on computer vision. pp. 1385--1392 (2013)

\bibitem{xu2019towards}
Xu, X., Ma, Y., Sun, W.: Towards real scene super-resolution with raw images.
  In: Proceedings of the IEEE/CVF Conference on Computer Vision and Pattern
  Recognition. pp. 1723--1731 (2019)

\bibitem{xue2019video}
Xue, T., Chen, B., Wu, J., Wei, D., Freeman, W.T.: Video enhancement with
  task-oriented flow. International Journal of Computer Vision
  \textbf{127}(8),  1106--1125 (2019)

\bibitem{yang2021real}
Yang, X., Xiang, W., Zeng, H., Zhang, L.: Real-world video super-resolution: A
  benchmark dataset and a decomposition based learning scheme. In: Proceedings
  of the IEEE/CVF International Conference on Computer Vision. pp. 4781--4790
  (2021)

\bibitem{yue2020supervised}
Yue, H., Cao, C., Liao, L., Chu, R., Yang, J.: Supervised raw video denoising
  with a benchmark dataset on dynamic scenes. In: Proceedings of the IEEE/CVF
  Conference on Computer Vision and Pattern Recognition. pp. 2301--2310 (2020)

\bibitem{zhang2019zoom}
Zhang, X., Chen, Q., Ng, R., Koltun, V.: Zoom to learn, learn to zoom. In:
  Proceedings of the IEEE/CVF Conference on Computer Vision and Pattern
  Recognition. pp. 3762--3770 (2019)

\bibitem{zhang2018image}
Zhang, Y., Li, K., Li, K., Wang, L., Zhong, B., Fu, Y.: Image super-resolution
  using very deep residual channel attention networks. In: Proceedings of the
  European conference on computer vision (ECCV). pp. 286--301 (2018)

\bibitem{zhou2021revisiting}
Zhou, K., Li, W., Lu, L., Han, X., Lu, J.: Revisiting temporal alignment for
  video restoration. arXiv preprint arXiv:2111.15288  (2021)

\end{thebibliography}
\end{document}